\documentclass[11pt]{article}

\usepackage[final]{acl}
\usepackage{times}
\usepackage{latexsym}

\usepackage[T1]{fontenc}

\usepackage[utf8]{inputenc}

\usepackage{microtype}

\usepackage{inconsolata}

\usepackage{graphicx}

\usepackage{amsmath}
\usepackage{amssymb}
\usepackage{xcolor}
\usepackage{booktabs}
\usepackage{multirow}
\usepackage{graphicx}
\usepackage{graphicx}
\title{CURA: Clinical Uncertainty Risk Alignment for Language Model–Based Risk Prediction}

\author{
  Sizhe Wang\textsuperscript{1} \quad Ziqi Xu\textsuperscript{1} \quad Claire Najjuuko\textsuperscript{1} \\ 
  \textbf{\quad Charles Alba\textsuperscript{1} \quad Chenyang Lu\textsuperscript{1}} \\
  \textsuperscript{1}Washington University in St. Louis \\
  \texttt{\{sizhew, ziqixu, c.najjuuko, alba, lu\}@wustl.edu}
}

\begin{document}
\maketitle
\begin{abstract}
Clinical language models (LMs) are increasingly applied to support clinical risk prediction from free-text notes, yet their uncertainty estimates often remain poorly calibrated and clinically unreliable.
In this work, we propose \underline{\textbf{C}}linical \underline{\textbf{U}}ncertainty \underline{\textbf{R}}isk \underline{\textbf{A}}lignment (\textbf{CURA}), 
a framework that aligns clinical LM-based risk estimates and uncertainty with both individual error likelihoods and cohort-level ambiguities.
CURA first fine-tunes domain-specific clinical LMs to obtain task-adapted patient embeddings, and then performs uncertainty fine-tuning of a multi-head classifier using a \textit{bi-level} uncertainty objective.
Specifically, an individual-level calibration term aligns predictive uncertainty with each patient’s likelihood of error, 
while a cohort-aware regularizer pulls risk estimates toward event rates in their local neighborhoods in the embedding space and places extra weight on ambiguous cohorts near the decision boundary.
We further show that this cohort-aware term can be interpreted as a cross-entropy loss with neighborhood-informed soft labels, providing a label-smoothing view of our method.
Extensive experiments on MIMIC-IV clinical risk prediction tasks across various clinical LMs show that CURA consistently improves calibration metrics without substantially compromising discrimination.
Further analysis illustrates that CURA reduces overconfident false reassurance and yields more trustworthy uncertainty estimates for downstream clinical decision support. 
Our code is available at: \url{https://github.com/sizhe04/CURA}.
\end{abstract}

\begin{figure}[t] 
    \centering
    \includegraphics[width=\linewidth]{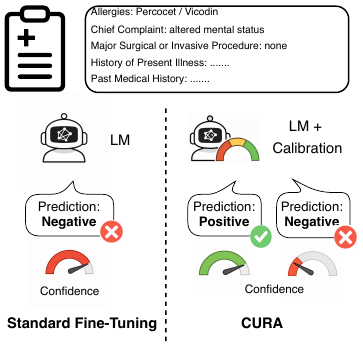}
    \caption{
    Comparison between standard fine-tuning and our proposed CURA framework in clinical risk prediction. 
    A vanilla fine-tuned clinical LM (left) may produce overconfident wrong predictions, while CURA (right) encourages high confidence only for correct decisions and assigns higher uncertainty to potentially erroneous predictions.
    }    
    \label{fig:pre_figure}
\end{figure}

\section{Introduction}
Clinical language models (LMs)~\citep{alsentzer2019publicly,huang2019clinicalbert,luo2022biogpt} have demonstrated remarkable capabilities in processing unstructured electronic health records (EHRs) to support clinical decision-making~\citep{jiang2023health,yang2022large}.
By fine-tuning on free-text clinical notes, these models can effectively extract patient representations to predict outcomes such as mortality risk and length of stay~\citep{harutyunyan2019multitask,johnson2023mimic}.
However, a key barrier to deploying such models in safety-critical settings is the reliability of their uncertainty estimates. 
While recent advancements have pushed the boundaries of discriminative performance~\citep{nori2023capabilities,singhal2025toward}, overconfident yet incorrect predictions still pose a direct danger to patient safety.

In clinical scenarios, reliable uncertainty quantification is crucial alongside prediction accuracy.
A well-calibrated model should act as a transparent partner, signaling high uncertainty for ambiguous or out-of-distribution cases to trigger expert review, while automating low-risk, high-confidence predictions.
However, although fine-tuning typically improves predictive performance, it often exacerbates the model’s tendency toward overconfidence~\citep{guo2017calibration}, leading to cases where the model is highly confident yet wrong on high-risk patients as shown in Figure~\ref{fig:pre_figure} (left).
To mitigate this issue, various calibration techniques have been proposed. 
General uncertainty estimation methods, such as Monte Carlo (MC) Dropout~\citep{gal2016dropout} and Deep Ensembles~\citep{lakshminarayanan2017simple} estimate uncertainty by aggregating predictions from multiple stochastic or independently trained models.
In parallel, recent works have explored calibration strategies tailored for generative  LLMs, often leveraging verbalized probabilities or prompting techniques~\citep{wang2024calibrating,leng2024taming,huang2025efficient,huang2025beyond}.

Nevertheless, directly applying these approaches in clinical settings remains challenging. 
First, general ensemble-based methods typically aggregate multiple predictions without leveraging the semantic structure in the representation space, adjust confidence scores only at the level of isolated samples, and may remain miscalibrated on borderline cohorts.
Second, many LLM-specific calibration methods rely on expert-written rationales, high-quality chain-of-thought (CoT) annotations, or strong teacher models to refine verbalized uncertainty~\citep{yang2024verbalized}. 
However, in routine clinical workflows, downstream tasks are typically framed as binary risk stratification with only outcome labels available, and ground-truth explanations are rarely recorded at scale. 
Moreover, for complex patient profiles, even state-of-the-art LLMs may produce clinical reasoning that is difficult to trust, especially if such explanations are used as supervision targets~\citep{nori2023capabilities}. 
These constraints highlight the need for uncertainty calibration methods that operate directly on learned representations and binary labels, without relying on explicit textual rationales.

To address these limitations, we propose \textbf{CURA} (Clinical Uncertainty Risk Alignment), a bi-level (individual- and cohort-level) uncertainty calibration framework built on top of fine-tuned clinical LMs for risk prediction.
As shown in Figure~\ref{fig:pre_figure} (right), CURA aims to reserve high confidence for correct predictions while assigning higher uncertainty to potentially erroneous decisions.
Specifically, our approach first fine-tunes a domain-specific clinical LM on free-text notes using standard cross-entropy to obtain task-adapted patient embeddings. 
On top of these frozen representations, we then train a multi-head classifier with a bi-level uncertainty objective that simultaneously aligns uncertainty at the individual and cohort levels.
Additionally, this objective can be implemented as a lightweight plug-in loss on top of standard clinical LM fine-tuning, without relying on external explanations or multiple inference passes.
Extensive experiments on MIMIC-IV clinical risk prediction tasks show that CURA consistently improves calibration metrics while maintaining—or slightly improving—discriminative performance, and further analysis demonstrates that CURA reshapes the uncertainty distribution, reduces false reassurance for high-risk patients, and provides greater practical utility for real-world clinical deployment.


To summarize, our contribution in this work is threefold:

\begin{itemize}
    \item We propose CURA, a bi-level uncertainty calibration framework that calibrates clinical LMs by jointly aligning individual predictive uncertainty with cohort-level risk via neighbors in the embedding space.
    
    \item We formulate CURA as a lightweight plug-in objective that requires no architectural changes and admits an interpretation as cross-entropy with neighborhood-informed soft labels.

    \item Extensive experiments and analysis show that CURA not only improves calibration metrics but also yields trustworthy uncertainty estimates for safe clinical triage.
\end{itemize}

\section{Related Work}

\paragraph{Clinical LMs for Risk Prediction}
Recent work has shown that domain-specific clinical LMs, from BERT-style encoders~\citep{alsentzer2019publicly,huang2019clinicalbert,lee2020biobert} to generative LLMs such as BioGPT~\citep{luo2022biogpt}, can substantially improve risk prediction from free-text notes and other electronic health record (EHR) data~\citep{singhal2022large,yang2022large,jiang2023health,alba2025foundational}.

However, the vast majority of these studies are primarily optimized and evaluated on discriminative metrics, with little attention to the calibration of the resulting risk estimates~\citep{harutyunyan2019multitask,zhu2024prompting,chen2025narrative}.
Our work distinguishes itself by proposing a framework that is designed to substantially improve the calibration of clinical LM–based risk prediction while maintaining, and in many cases slightly improving, discriminative performance.

\paragraph{Uncertainty Calibration for LLMs}
Uncertainty quantification is fundamental for reliable AI deployment in safety-critical domains such as healthcare~\citep{kendall2017uncertainties,abdar2021review,gawlikowski2023survey,xu2025incorporating}.
Classic approaches approximate Bayesian inference through techniques such as MC dropout and deep ensembles, while post-hoc calibration methods adjust confidence scores using temperature scaling or related transformations~\citep{guo2017calibration,kull2019beyond}.
In the era of LLMs, research has shifted toward eliciting uncertainty via verbalized confidence, self-evaluation, or semantic consistency~\citep{lin2022teaching,tian2023just,kadavath2022language,melo2024deep,kapoor2024large}.
Beyond these generation-centric approaches, LLM-based reward models adopt multi-head ensembles on a shared backbone to obtain diverse uncertainty estimates at low computational cost~\citep{liang2022reward,duan2025efficient}.
While promising for open-ended question answering and reasoning benchmarks, these generation-centric methods' calibration behavior in classification tasks remains largely unexplored~\citep{liu2025uncertainty} and is susceptible to overconfidence on distribution-shifted inputs~\citep{zhou2023navigating,zulfiqar2025uncertainty}.
Crucially, most prior uncertainty quantification techniques operate on individual samples in isolation, neglecting the structural information carried by similar examples. 
Our work bridges this gap by explicitly aligning predictive uncertainty with both individual error rates and cohort neighborhood risks in the embedding space, in a white-box setting where the clinical LM and prediction heads are fully accessible, rather than via prompt-based black-box access.






\begin{figure*}[t] 
    \centering
    \includegraphics[width=\textwidth,
  trim=0 0 0 0, 
  clip]{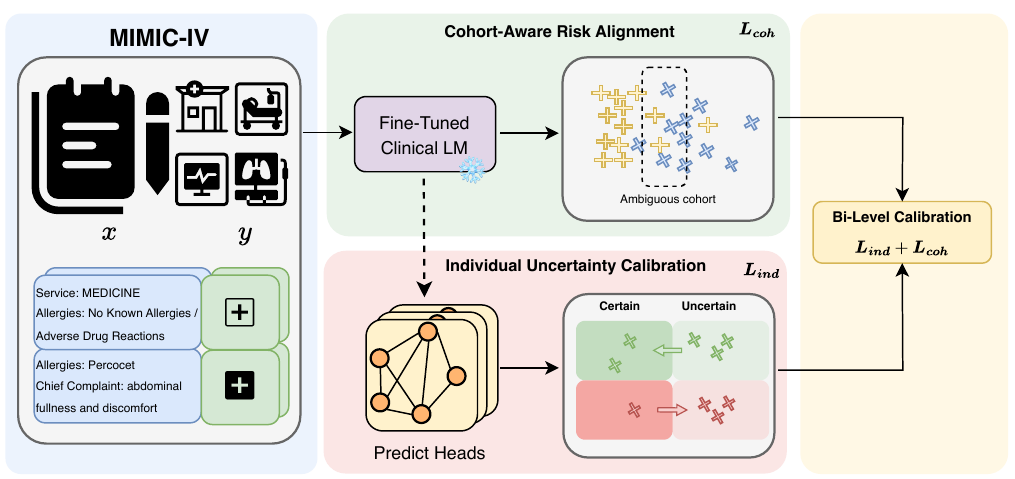}
    \caption{
    Overview of the CURA pipeline. 
    The clinical LM is fine-tuned on MIMIC-IV notes to produce task-adapted embeddings for a multi-head classifier, trained with a base loss $L_{base}$ plus a bi-level uncertainty calibration: the individual term $L_{ind}$ calibrates patient-level uncertainty, and the cohort-aware term $L_{coh}$ aligns predictions with neighborhood risks in the embedding space, emphasizing ambiguous cohorts.    
    }

    \label{fig:pipeline}
\end{figure*}

\section{Methodology}
In this section, we elaborate on the Clinical Uncertainty Risk Alignment (CURA) framework, depicted in Figure~\ref{fig:pipeline}, which consists of two primary components:
Section~\ref{sec:clinical-LM-fine-tuning} describes fine-tuning domain-specific clinical LMs, and
Section~\ref{sec:uncertainty-fine-tuning} presents the uncertainty fine-tuning.

\subsection{Clinical LM Fine-Tuning}
\label{sec:clinical-LM-fine-tuning}
We address clinical risk prediction tasks using a dataset $\mathcal{D} = \{(x_i, y_i)\}$, where each input $x_i$ consists of a de-identified clinical note, and the target $y_i \in \{0,1\}$ serves as a binary indicator for the downstream task.
We initialize our framework with a domain-specific clinical LM $\pi_{0}$ and attach a linear classifier on top. 

Let $\pi_{\theta}$ denote the LM parameterized by $\theta$, and let $p_{\theta}(x)$ be the predicted positive-class probability produced by the linear classifier on top of $\pi_{\theta}$. 
We jointly optimize the parameters of the language model $\pi_{\theta}$ and the linear classifier by minimizing a class-weighted binary cross-entropy~\citep{xie1989logit} loss $\mathrm{CE}(p_\theta(x), y)$ over $\mathcal{D}$.
After fine-tuning, we freeze the updated LM $\pi_{\theta}$ and use it as a feature extractor to obtain fixed-dimensional embeddings $e_i = \pi_{\theta}(x_i)$ for all samples. 
These frozen embeddings serve as the input for the subsequent uncertainty fine-tuning step.

\subsection{Uncertainty Fine-Tuning}
\label{sec:uncertainty-fine-tuning}
Our proposed objective aligns three complementary aspects: 
(1) $L_{base}$ ensures predictive accuracy against ground-truth labels (Section~\ref{sec:base-loss}); 
(2) $L_{ind}$ aligns the model’s internal uncertainty with its own estimated error proxy (Section~\ref{sec:calibration}); and
(3) $L_{coh}$ regularizes predictions via local consistency with cohort risks (Section~\ref{sec:cohort}).
Throughout, we use the term \textit{bi-level} to refer to these individual- and cohort-level components of uncertainty alignment, rather than a nested bilevel optimization problem.


\subsubsection{Classifier and Base Risk Loss}
\label{sec:base-loss}
We construct a multi-head framework with an ensemble of $M$ independent, randomly initialized lightweight MLP heads on top of the frozen embeddings $\{e_i\}$.
At inference, we average predictions across all $M$ heads.
This architecture encourages diverse risk prediction and stabilizes calibration while sharing a single backbone for minimal inference cost~\citep{duan2025efficient}.
Given a patient note $x_i$ with embedding $e_i = \pi_{\theta}(x_i)$, 
the ensemble average prediction is $\bar{p}(x_i) = \frac{1}{M} \sum_{m=1}^M p_m(e_i)$,
and the base objective optimizes the expected cross-entropy between this mean prediction and the target $y$:
\begin{equation}
\label{base_loss}
    L_{base} = \mathbb{E}_{(x_i, y_i)} \big[ \mathrm{CE}(\bar{p}(x_i), y_i) \big],
\end{equation}
where $\mathrm{CE}(\cdot,\cdot)$ is the same class-weighted loss used in the clinical LM fine-tuning step. 
This base loss provides the discriminative backbone for our uncertainty fine-tuning.

\subsubsection{Individual Uncertainty Calibration}
\label{sec:calibration}

While $L_{base}$ ensures accurate risk estimation, it does not explicitly constrain how the model’s confidence relates to its errors, so its predicted probability can still be poorly calibrated.
Motivated by prior work on uncertainty-aware fine-tuning for LLMs~\citep{krishnan2024enhancing}, which is grounded in decision theory~\citep{murphy2012machine} and couples token-level predictive entropy with correctness in inference, we introduce an analogous individual-level calibration term for binary risk prediction.
For a patient note $x$, we define the correctness probability $a(x) = y\,\bar{p}(x) + (1-y)(1-\bar{p}(x))$, representing the confidence assigned to the ground-truth label.
Based on information-theoretic views of predictive uncertainty, where entropy is used as a  measure of distributional uncertainty~\citep{shannon1948mathematical,houlsby2011bayesian}, we quantify the model’s uncertainty via the entropy of the ensemble-averaged prediction:
\begin{equation}
\label{eq:entropy}
\begin{aligned}
    H(x) = &-\Bigl[\,\bar{p}(x)\log \bar{p}(x) \\
&+ \bigl(1-\bar{p}(x)\bigr)\log\bigl(1-\bar{p}(x)\bigr)\Bigr].
\end{aligned}
\end{equation}
We normalize this entropy by its maximum possible value and obtain a scalar uncertainty score $u(x) = H(x) / H_{\max} \in [0,1]$, and align it with the error proxy $(1-a(x))$ by minimizing the individual-level calibration loss:

\begin{equation}
\label{eq:L-indiv}
\begin{aligned}
    L_{ind} = \lambda_{ind}\,\mathbb{E}_{(x,y)}\Bigl[&-(1-a(x))\log u(x) \\
                        &- a(x)\log\bigl(1-u(x)\bigr)\Bigr],
\end{aligned}
\end{equation}
where $\lambda_{ind}$ is a scaling hyperparameter.
Minimizing $L_{ind}$ aligns the model's uncertainty with its error rate at the individual patient level:
confident and correct predictions are rewarded with small loss, whereas confident mistakes incur a large penalty.
Notably, the $L_{base}$ in Equation~\ref{base_loss} acts as an anchor for discrimination, preventing the degenerate solution where the model outputs uniform probabilities to minimize $L_{ind}$ trivially.
Furthermore, minimizing $L_{ind}$ effectively reshapes the probability distribution, penalizing high confidence specifically on samples where the model is prone to error.
In all subsequent analyses, we use the same normalized entropy $u(x)$ computed from each method’s final predicted probability as the scalar uncertainty score, so that all uncertainty-based comparisons are made under a common definition.


\subsubsection{Cohort-Aware Risk Alignment}
\label{sec:cohort}
To extend risk alignment from the individual to the group level, we propose a \textit{cohort-aware} term that ensures clinically similar patients receive consistent risk estimates.
For each patient note $x_i$ with embedding $e_i$, we retrieve its $K$ nearest neighbors, denoted as $\mathcal{N}_K(e_i)$, among all labeled training samples and compute the  event rate in this local cohort:
\begin{equation}
    q(x_i) = \frac{1}{K}\sum_{j\in\mathcal{N}_K(e_i)} y_j.
    \label{eq:q-cohort}
\end{equation}
We interpret $q(x_i)$ as the \textit{neighborhood risk}, representing the inherent risk for patients with similar clinical presentations. 
To quantify cohort ambiguity, we compute the normalized entropy $\hat{H}(q(x_i))$, which increases when the cohort exhibits a mixed outcome distribution—a characteristic of borderline cases.
We leverage this cohort information to regularize the model via a cohort-aware loss:
\begin{equation}
\label{eq:L-coh}
    L_{coh} = \mathbb{E}_{x_i}\bigl[w(x_i)\,\mathrm{CE}\bigl(\bar p(x_i), q(x_i)\bigr)\bigr],
\end{equation}
where the adaptive weight $w(x_i) = \lambda_{coh}\,\hat{H}\bigl(q(x_i)\bigr)$ is controlled by the hyperparameter $\lambda_{coh}$.
This objective aligns $\bar{p}(x_i)$ with the neighborhood risk $q(x_i)$ and assigns larger training weights to cohorts with high neighborhood entropy.

Overall, our uncertainty fine-tuning objective is:
\begin{equation}
\label{eq:L-total}
    L_{total} = L_{base} + L_{ind} + L_{coh}.
\end{equation}
Here $L_{base}$ provides standard risk supervision for the multi-head classifier, while $L_{ind}$ and $L_{coh}$ together form the bi-level uncertainty regularizer illustrated in Figure~\ref{fig:pipeline}.
In Appendix~\ref{app:cohort-softlabel}, we further show that minimizing $L_{base} + L_{coh}$ is equivalent to optimizing a single cross-entropy loss with a cohort-aware soft label, which can be viewed as a data-dependent label-smoothing scheme whose target interpolates between the ground-truth label and the neighborhood risk.
This soft-label view highlights that cohorts with high neighborhood entropy pull the target toward their local event rate and thus receive stronger regularization, while in practice $L_{total}$ can be used as a drop-in training objective on top of existing clinical LMs without modifying their inference pipeline.
Because $L_{ind}$ and $L_{coh}$ are computed from model probabilities and pre-computed neighborhoods, CURA does not introduce additional backbone parameters or extra forward passes beyond the shared single-model pipeline. 
An empirical runtime comparison is provided in Appendix~\ref{app:runtime}.


\begin{table*}[t!]
\centering
\small
\setlength{\tabcolsep}{5.5pt}
\renewcommand{\arraystretch}{1.2}
\begin{tabular}{llccccc}
\toprule
\textbf{Task} & \textbf{Method} & \textbf{AUROC} $\uparrow$ & \textbf{AUPRC} $\uparrow$ & \textbf{Brier} $\downarrow$ & \textbf{NLL} $\downarrow$ & \textbf{AURC} $\downarrow$ \\
\midrule
\multirow{4}{*}{\textbf{7-day Mortality}} 
& Baseline & 0.852 (0.025) & 0.127 (0.015) & 0.032 (0.005) & 0.120 (0.015) & 0.008 (0.001) \\
& Deep Ensemble & 0.856 (0.022) & 0.125 (0.010) & 0.029 (0.004) & 0.110 (0.009) & 0.007 (0.000) \\
& MC Dropout & 0.862 (0.019) & 0.130 (0.008) & 0.034 (0.006) & 0.127 (0.017) & 0.009 (0.002) \\
& \textbf{CURA (Ours)} & \textbf{0.892} (0.010) & \textbf{0.160} (0.010) & \textbf{0.015} (0.003) & \textbf{0.075} (0.010) & \textbf{0.002} (0.000) \\
\midrule
\multirow{4}{*}{\textbf{30-day Mortality}} 
& Baseline & 0.881 (0.009) & 0.270 (0.009) & 0.064 (0.003) & 0.231 (0.004) & 0.024 (0.003) \\
& Deep Ensemble & 0.869 (0.008) & 0.256 (0.011) & 0.060 (0.001) & 0.215 (0.004) & 0.020 (0.002) \\
& MC Dropout & 0.880 (0.014) & 0.268 (0.014) & 0.063 (0.002) & 0.230 (0.007) & 0.024 (0.003) \\
& \textbf{CURA (Ours)} & \textbf{0.890} (0.004) & \textbf{0.280} (0.012) & \textbf{0.038} (0.001) & \textbf{0.146} (0.003) & \textbf{0.009} (0.001) \\
\midrule
\multirow{4}{*}{\textbf{In-hospital Mortality}} 
& Baseline & 0.621 (0.011) & 0.027 (0.002) & 0.044 (0.004) & 0.175 (0.013) & 0.015 (0.001) \\
& Deep Ensemble & 0.632 (0.010) & 0.028 (0.002) & 0.044 (0.004) & 0.171 (0.010) & 0.015 (0.001) \\
& MC Dropout & 0.629 (0.013) & 0.028 (0.002) & 0.046 (0.005) & 0.174 (0.013) & 0.015 (0.001) \\
& \textbf{CURA (Ours)} & \textbf{0.641} (0.009) & \textbf{0.029} (0.002) & \textbf{0.029} (0.004) & \textbf{0.124} (0.010) & \textbf{0.011} (0.000) \\
\midrule
\multirow{4}{*}{\textbf{ICU Stay - 1}} 
& Baseline & 0.579 (0.026) & \textbf{0.035} (0.003) & 0.105 (0.042) & 0.350 (0.108) & 0.064 (0.053) \\
& Deep Ensemble & 0.579 (0.024) & \textbf{0.035} (0.002) & 0.104 (0.043) & 0.345 (0.110) & 0.062 (0.052) \\
& MC Dropout & 0.580 (0.027) & \textbf{0.035} (0.003) & 0.106 (0.043) & 0.350 (0.113) & 0.066 (0.056) \\
& \textbf{CURA (Ours)} & \textbf{0.584} (0.029) & \textbf{0.035} (0.003) & \textbf{0.085} (0.044) & \textbf{0.288} (0.104) & \textbf{0.040} (0.030) \\
\midrule
\multirow{4}{*}{\textbf{Early Discharge - 12}} 
& Baseline & 0.587 (0.011) & 0.027 (0.012) & 0.018 (0.002) & 0.085 (0.006) & 0.007 (0.001) \\
& Deep Ensemble & 0.586 (0.012) & 0.028 (0.012) & 0.017 (0.001) & 0.085 (0.005) & 0.007 (0.000) \\
& MC Dropout & 0.588 (0.013) & 0.028 (0.012) & 0.018 (0.002) & 0.083 (0.008) & 0.007 (0.000) \\
& \textbf{CURA (Ours)} & \textbf{0.594} (0.017) & \textbf{0.031} (0.014) & \textbf{0.010} (0.001) & \textbf{0.056} (0.003) & \textbf{0.006} (0.000) \\
\bottomrule
\end{tabular}
\caption{Performance of \textbf{BioGPT} on five clinical risk prediction tasks. Results are reported as mean (standard deviation) across five-fold cross-validation. 
``ICU Stay - 1'' refers to ICU stays longer than one day, and ``Early Discharge - 12'' refers to early discharge within 12 hours.}
\label{tab:biogpt_results}
\end{table*}

\section{Experiment}
\label{sec:experiment}
\subsection{MIMIC-IV Dataset}
\label{sec:MIMIC}


We utilize de-identified clinical notes from the MIMIC-IV database~\citep{johnson2023mimic} to predict five distinct risk stratification tasks.
For each task, the input $x$ is a single free-text note and the output $y \in \{0,1\}$ indicates a pre-defined adverse outcome within a fixed horizon.
These tasks align with benchmarks in prior clinical LM research~\citep{xue2022perioperative,luo2022association,alba2025foundational} and cover diverse clinical scenarios, ranging from short-term to medium-term prognosis.
Specifically, we evaluate on:
(1) 7-day mortality,
(2) 30-day mortality,
(3) early discharge within 12 hours,
(4) in-hospital mortality, and
(5) ICU stay longer than one day.
Further details regarding the task definitions are provided in Appendix~\ref{app:MIMIC}.

\subsection{Experiment Settings}

\paragraph{Clinical Language Models}
We utilize three domain-specific language model backbones: BioGPT \citep{luo2022biogpt}, BioClinicalBERT \citep{alsentzer2019publicly}, and ClinicalBERT \citep{huang2019clinicalbert}.
For brevity, we present results for BioGPT in the main experiment, and defer the BioClinicalBERT and ClinicalBERT results to Appendix~\ref{app:BioClinicalBERT-ClinicalBERT-result}. 

\paragraph{Baselines}
Following prior work on uncertainty quantification for clinical outcome prediction \citep{chen2025uncertainty}, we include two widely used architecture-agnostic white-box baselines: Deep Ensembles and MC dropout.
In addition, we also consider an internal baseline that uses the same multi-head architecture as CURA but optimizes only the class-weighted cross-entropy in the uncertainty fine-tuning step.
All baselines employ the same fine-tuned backbone models as our approach, so that any performance differences stem only from the choice of uncertainty mechanism.
We also report a runtime comparison under identical hardware and early-stopping settings in Appendix~\ref{app:runtime} to evaluate the practical overhead of each uncertainty method.
Further implementation details are provided in Appendix~\ref{app:experiment-setup}.

\paragraph{Evaluation Metrics}
We evaluate model performance using two categories of metrics: (1) Discrimination: Area Under the ROC Curve (AUROC) and Area Under the Precision-Recall Curve (AUPRC); and (2) Calibration \& Uncertainty: Brier score, Negative Log-Likelihood (NLL), and Area Under the Risk-Coverage curve (AURC).
All reported results represent the mean and standard deviation across five cross-validation folds.


\subsection{Experimental Results}
\label{sec:main_results}
In this section, we compare our overall CURA method with the baseline and the two aforementioned uncertainty quantification strategies.
As shown in Table~\ref{tab:biogpt_results}, CURA achieves modest but stable improvements in discrimination performance (AUROC and AUPRC) relative to the baseline, indicating that our method does not sacrifice predictive accuracy. 
While Deep Ensembles and MC Dropout remain competitive in discriminative tasks, their contribution to calibration appears less pronounced in this setting.
These methods yield only marginal reductions in Brier score, NLL, and AURC, and in some instances even slightly degrade these metrics.
Comparatively, CURA exhibits stronger calibration capabilities, yielding consistent reductions across all three calibration metrics on all five tasks.
We attribute this effectiveness to the synergistic effect of our individual and cohort-level alignment, which allows the model to better distinguish between reliable and ambiguous predictions.
Overall, these results suggest that our bi-level uncertainty fine-tuning objective is particularly effective at reshaping the probability outputs of clinical LMs into reliable risk estimates, while retaining competitive discriminative performance.
Similar robust trends are observed for BioClinicalBERT and ClinicalBERT, as detailed in Appendix~\ref{app:BioClinicalBERT-ClinicalBERT-result}.

\begin{figure*}[t]
    \centering
    \includegraphics[width=\textwidth]{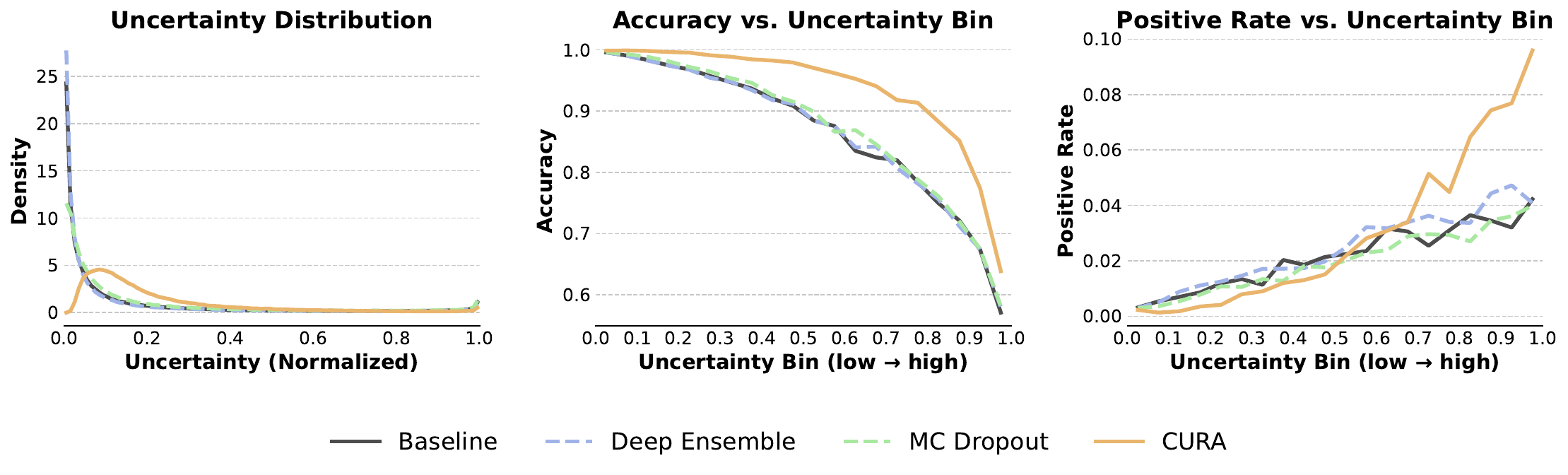}
    \caption{\textbf{Left:} Distribution of normalized uncertainty estimates, 
    \textbf{Middle:} prediction accuracy within each uncertainty bin at a fixed decision threshold ($0.5$) on predicted risk, and 
    \textbf{Right:} positive rate within each uncertainty bin for the 7-day mortality task with BioGPT.
    }
    \label{fig:threefigures}
\end{figure*}

\begin{table}[t]
\centering
\small
\setlength{\tabcolsep}{4pt}
\renewcommand{\arraystretch}{1.2}
\begin{tabular}{lccccc}
\toprule
\textbf{Method} & \textbf{AUROC} & \textbf{AUPRC} & \textbf{Brier} & \textbf{NLL} & \textbf{AURC} \\
\midrule
\multicolumn{6}{l}{\textit{7-day Mortality}} \\
Baseline & 0.852 & 0.127 & 0.032 & 0.120 & 0.008 \\
w/ $L_{\mathrm{ind}}$ & \textbf{0.892} & \textbf{0.160} & 0.019 & 0.077 & 0.003 \\
w/ $L_{\mathrm{coh}}$ & 0.851 & 0.124 & 0.023 & 0.118 & 0.007 \\
\textbf{CURA} & \textbf{0.892} & \textbf{0.160} & \textbf{0.015} & \textbf{0.075} & \textbf{0.002} \\
\midrule
\multicolumn{6}{l}{\textit{In-hospital Mortality}} \\
Baseline & 0.621 & 0.027 & 0.044 & 0.175 & 0.015 \\
w/ $L_{\mathrm{ind}}$ & \textbf{0.641} & \textbf{0.029} & 0.035 & 0.128 & 0.012 \\
w/ $L_{\mathrm{coh}}$ & 0.625 & 0.028 & 0.040 & 0.166 & 0.014 \\
\textbf{CURA} & \textbf{0.641} & \textbf{0.029} & \textbf{0.029} & \textbf{0.124} & \textbf{0.011} \\
\bottomrule
\end{tabular}
\caption{Ablation study on \textbf{BioGPT} for high-stakes mortality tasks. We isolate the contributions of individual calibration $L_{ind}$ and cohort alignment $L_{coh}$. Standard deviations are omitted for brevity.}
\label{tab:ablation_main}
\end{table}

\subsection{Ablation Study}
\label{sec:ablation}
We perform ablation studies on BioGPT to isolate the contributions of the two proposed uncertainty  components in our objective.
Starting from an internal baseline that uses only class-weighted cross-entropy, we incrementally add 
(i) the individual calibration term $L_{ind}$, 
(ii) the cohort-aware risk alignment term $L_{coh}$, and 
(iii) both terms, which together yield the full CURA objective. 
In the main text we report ablations on two clinically high-stakes tasks, 7-day mortality and in-hospital mortality, and defer the remaining three tasks to Appendix~\ref{app:ablation_full}. As shown in Table~\ref{tab:ablation_main}, across both tasks, adding either $L_{ind}$ or $L_{coh}$ consistently improves calibration metrics such as Brier score, NLL, and AURC, while leaving AUROC and AUPRC essentially unchanged or slightly improved. 
Combining both terms yields the best overall calibration performance without sacrificing discriminative accuracy.
Additionally, we evaluate the sensitivity of CURA to the loss weights $\lambda_{ind}$ and $\lambda_{coh}$ on two high-stakes mortality tasks, with detailed results provided in Appendix~\ref{app:loss_weight_sensitivity}.

\section{Uncertainty Analysis for Clinical Triage}
\label{sec:further-analysis}

In this section, we analyze how CURA reshapes uncertainty in deployment-relevant clinical triage, focusing on the BioGPT backbone and the 7-day mortality task where calibration gains are most pronounced.
Unless otherwise noted, per-patient uncertainty is measured by the normalized entropy score $u(x)$ from Section~\ref{sec:calibration}, and all results are averaged over five cross-validation folds.

\begin{figure*}[t]
    \centering
    \includegraphics[width=\textwidth]{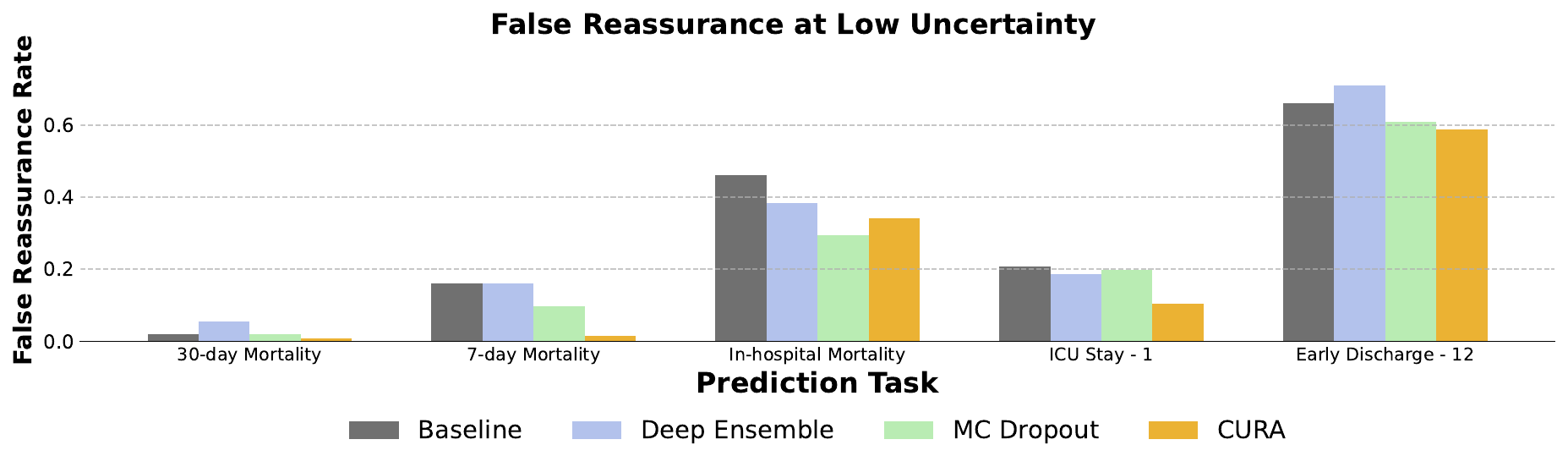}
    \caption{False reassurance rates at low uncertainty across five prediction tasks with BioGPT. 
    Lower values indicate fewer high-risk patients being confidently misclassified as safe.}
    \label{fig:false_reassurance}
\end{figure*}

\begin{figure} [t] 
    \centering
    \includegraphics[width=\linewidth]{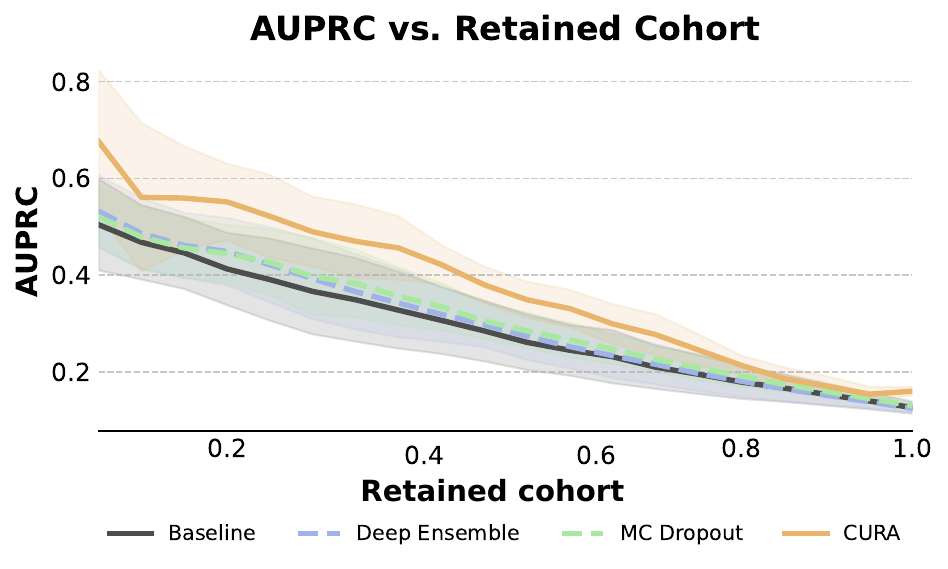}
    \caption{
    AUPRC on the retained cohort as a function of the retained fraction.
    Patients are sorted by uncertainty, and only the lowest-uncertainty fraction is retained for prediction.
    }    
    \label{fig:AUROC&AUPRC}
\end{figure}

\subsection{Uncertainty-driven Risk Stratification}
\label{sec:fa-risk-strat}

Figure~\ref{fig:threefigures} illustrates the interplay between predictive uncertainty, accuracy, and clinical risk.
The left panel displays the distribution of uncertainty $u(x)$, showing that the baseline, deep ensembles, and MC dropout all exhibit severe overconfidence, with probability mass heavily concentrated near zero uncertainty. 
In contrast, CURA produces a dispersed uncertainty distribution, effectively shifting probability mass from the overconfident spike to moderate-to-high uncertainty regions.

The middle panel reports classification accuracy within each uncertainty bin.
CURA generally achieves higher bin-wise accuracy than other methods, particularly in the moderate-to-high uncertainty range. 
Combined with the distribution shift, this indicates that CURA both reduces the number of predictions that the model considers nearly certain and makes each uncertainty level more trustworthy.

Furthermore, the right panel plots the positive rate within each uncertainty bin.
All methods show a correlation between uncertainty and risk, but CURA demonstrates the strongest alignment: low-uncertainty bins are dominated by negative cases, whereas high-uncertainty bins are substantially enriched with positive cases.
Clinically, this pattern is desirable, since low-uncertainty predictions form a safer pool of patients for automatic triage, whereas high-uncertainty predictions correspond to high-risk, ambiguous cases that should be escalated for doctor review.

Overall, our method mitigates overconfidence and concentrates clinical risk in the high-uncertainty region, yielding uncertainty estimates that are better aligned with risk stratification.



\subsection{Selective Deployment and Triage Workload}
\label{sec:fa-selective}


We evaluate the utility of calibrated uncertainties in supporting selective deployment, a paradigm where the model automates predictions for high-confidence patients while deferring uncertain cases to clinicians.
Figure~\ref{fig:AUROC&AUPRC} illustrates the AUPRC on the retained cohort, constructed by preserving only the patients with the lowest uncertainty and progressively increasing the retention fraction.
CURA consistently outperforms the baselines across the majority of retention levels, indicating better discriminative performance when employed as a decision-support tool for its most confident predictions.

\begin{figure} 
    \centering
    \includegraphics[width=\linewidth]{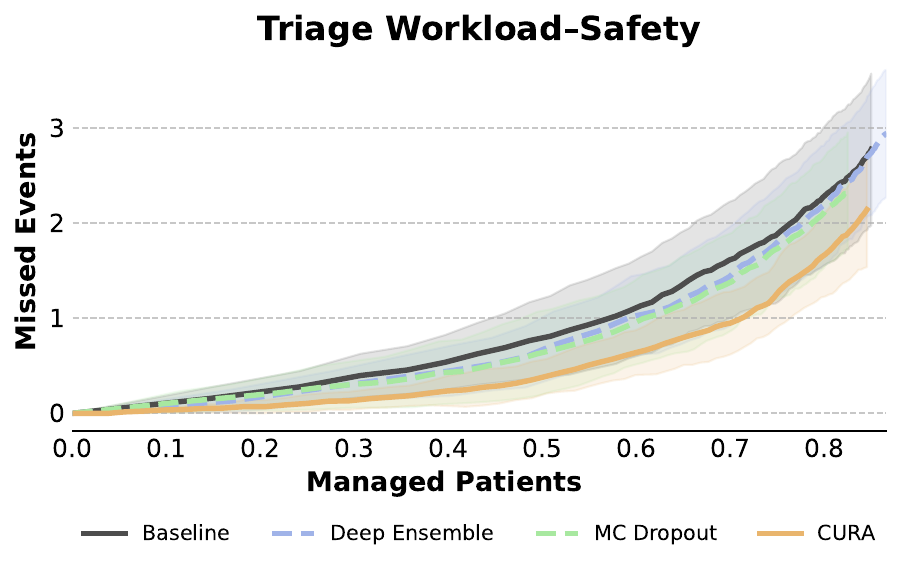}
    \caption{Triage workload--safety trade-off. 
    The curves plot the expected number of missed positive events per 1,000 patients against the fraction of patients automatically managed by the model.}
    \label{fig:workload-safety}
\end{figure}

To quantify the clinical implications, Figure~\ref{fig:workload-safety} depicts the triage workload--safety trade-off.
In this analysis, patients are prioritized by uncertainty, with the lowest-uncertainty fraction automatically managed by the model (x-axis); the y-axis reports the expected number of missed positive events per 1,000 patients within this automated group.
Across most of the operating range, our method yields fewer missed events than comparison methods for the same level of workload reduction.
Equivalently, for a fixed safety threshold, CURA allows a larger proportion of patients to be screened automatically.
These results suggest that CURA effectively enhances automation efficiency in high-stakes mortality prediction without compromising patient safety.

\subsection{Reducing False Reassurance}
\label{sec:fa-false-reassurance}
We consider a safety metric termed the \emph{false reassurance rate}.
Specifically, we define a safe region in which both model uncertainty $u(x)<\tau$ and predicted risk $\bar{p}(x)<\tau$, with $\tau=0.1$.
Additional results using $\tau = 0.05$ and $\tau = 0.15$ are provided in Appendix~\ref{app:false-reassurance-results}, demonstrating consistent trends.
The false reassurance rate is the fraction of truly positive patients whose predictions fall into this safe region, capturing a critical clinical failure mode: high-risk patients who are confidently misclassified as low-risk. 
A detailed definition is provided in Appendix~\ref{app:false-reassurance}.

Figure~\ref{fig:false_reassurance} shows the false reassurance rate across all five clinical tasks.
CURA generally achieves the lowest rate in four out of five tasks, with substantial reductions in 7-day and 30-day mortality prediction. 
It remains competitive on in-hospital mortality, where MC Dropout achieves a marginally lower value. 
These results indicate that our method reduces the fraction of high-risk patients who might be given misleadingly reassuring predictions.


\section{Conclusion}
In this work, we propose CURA, a bi-level uncertainty calibration framework that calibrates clinical LMs by jointly optimizing individual predictive uncertainty and cohort-level local consistency. 
By grounding risk estimates in patient-embedding neighborhoods, CURA mitigates overconfidence, especially on ambiguous cases.
Extensive experiments on MIMIC-IV demonstrate that our method consistently improves calibration metrics while largely maintaining competitive discriminative performance. 
Moreover, our method substantially lowers false reassurance rates and yields uncertainty estimates that better support clinical triage.

\section*{Limitations}
Despite the promising results of our approach, it has several limitations. 
Our current framework relies on discriminative LMs providing scalar risk scores without textual explanations, and we do not evaluate closed-source API models, largely because the MIMIC-IV data use agreement prohibits sharing patient data with third-party platforms. 
Our experiments are restricted to the retrospective analysis of single-center clinical notes. 
Future work will explore integrating CURA with generative models to offer interpretable reasoning and extending the framework to multi-modal, longitudinal settings and broader patient populations.

\section*{Ethical Considerations}
Digital health datasets such as MIMIC reflect systemic health disparities experienced by minority populations, which can be inadvertently propagated by AI models. Prior work has documented such disparities in MIMIC-III and other electronic health record datasets in the United States~\citep{roosli2022peeking,tripathi2020fairness,wang2024assessing}. Because we build on clinical language models pre-trained on MIMIC-III and evaluate on MIMIC-IV, our findings should be interpreted in light of these fairness concerns, and our framework does not directly mitigate underlying structural biases in the data.

\section*{Acknowledgments}
This work was supported in part by the Fullgraf Foundation and the AI for Health Institute at Washington University in St. Louis. 
Charles Alba was partially supported by the National University of Singapore Development Grant and the Danforth Scholarship at Washington University in St. Louis. 
Claire Najjuuko was partially supported by the NIH Researcher Resilience Training Grant (R25MH118935-01). 
The content is solely the responsibility of the authors and does not necessarily represent the official views of the funding agencies.

\bibliography{custom}

\appendix


\section{MIMIC-IV Dataset and Task Definitions}
\label{app:MIMIC}
In this section, we describe the MIMIC-IV dataset used in our experiments and provide detailed definitions of the five clinical risk prediction tasks.

\subsection{Task Definitions}
\paragraph{7-day mortality}
We consider the cohort of adult inpatients with at least one discharge summary. 
This task targets imminent post-discharge deterioration. 
The label is assigned as positive if the patient dies within 7 days after hospital discharge, with a positive rate of 0.88\%.

\paragraph{30-day mortality}
This task utilizes the same cohort as the 7-day mortality task but extends the prediction window to assess medium-term prognosis. 
The label is positive if the patient dies within 30 days after hospital discharge, and negative otherwise. 
The positive rate is 2.87\%.

\paragraph{Early discharge within 12 hours}
This task identifies admissions suitable for rapid triage or those with short, uncomplicated stays. 
The cohort is restricted to patients who were discharged alive (excluding in-hospital mortality). 
The label is positive if the discharge occurs within 12 hours of admission, and negative if the stay exceeds 12 hours. 
The positive rate is 0.74\%.

\paragraph{In-hospital mortality}
For this task, the label is positive if the patient dies before hospital discharge and negative otherwise.
It corresponds to the standard in-hospital mortality flag in MIMIC and is widely used as a benchmark for clinical risk prediction models. The positive rate is 1.71\%.

\paragraph{ICU stay longer than one day}
We consider encounters that include an ICU admission record. 
The label is positive if the patient spends more than 24 hours in the ICU during the index stay (summed across all ICU transfers within the same admission), and negative otherwise. 
This task reflects early critical-care resource utilization and is relevant for planning ICU staffing and bed management. 
The positive rate is 2.50\%.

After preprocessing, the two post-discharge mortality tasks (7-day and 30-day) contain approximately 323K encounters. 
For the remaining three tasks (early discharge within 12 hours, in-hospital mortality, and ICU stay longer than one day), the processed cohort consists of approximately 142K encounters.
Note that for the early discharge task, we exclude encounters resulting in in-hospital death to decouple mortality outcomes from resource utilization metrics.

\subsection{Data Governance and De-identification}
All experiments use the MIMIC-IV database of electronic health records and clinical notes~\citep{johnson2023mimic}, which is released under the PhysioNet Credentialed Health Data License 1.5.0~\citep{mimiciv_v2_2}. Access to the dataset requires completion of human-subjects protection training and acceptance of the MIMIC data use agreement. A credentialed author on our team fulfilled these requirements and accessed the data on behalf of the project, and all analyses complied with the data use agreement.

The MIMIC-IV notes were de-identified by the dataset curators in accordance with the HIPAA Safe Harbor provision. In particular, multiple categories of protected health information (e.g., personal names, locations, serial numbers, and ages) were removed or replaced with surrogate identifiers, and calendar dates were systematically perturbed to reduce the risk of re-identification~\citep{johnson2023mimic}. We treat the resulting corpus as de-identified and did not attempt to recover any original identifiers.

\subsection{Clinical Notes and Preprocessing}
We focus on discharge summaries from MIMIC-IV, which are written by clinicians after hospital discharge to summarize the patient’s hospital course, treatments, procedures, and follow-up plans. Because these notes were heavily de-identified to meet HIPAA’s Safe Harbor standard, many entities identified as protected health information were removed and replaced with placeholder tokens such as ``\_\_\_''. As a result, the raw text often contains long sequences of underscores and other artifacts.

To obtain a cleaner corpus for modeling, we adapt preprocessing pipelines from prior work on clinical language models~\citep{huang2019clinicalbert,alsentzer2019publicly,alba2025foundational}. Specifically, we remove excessive placeholder tokens (e.g., long runs of underscores), normalize whitespace, and discard non-printable or unrecognized characters. We do not attempt to reconstruct protected health information or to enrich the notes with external structured data.

\subsection{Exclusion Criteria}
We include adult inpatient encounters with at least one discharge summary and with sufficient follow-up information to define the outcome labels described below. Patients without any recorded discharge note are excluded. We further remove encounters whose discharge note is timestamped after or during the recorded time of in-hospital death. Manual inspection of a small sample of such cases indicates that these notes primarily document the death itself rather than the clinical course preceding it. Excluding these encounters avoids potential label leakage from the text into the prediction target and yields a more realistic evaluation setting.


\section{Additional Experimental Results on Different Models}
\label{app:BioClinicalBERT-ClinicalBERT-result}
In this section, we provide a detailed performance analysis using two additional clinical language model backbones: BioClinicalBERT and ClinicalBERT. 
These experiments aim to verify the generalizability of the CURA framework across different architectures.

For BioClinicalBERT, the results in Table~\ref{tab:bioclinicalbert_results} exhibit trends that are highly consistent with our main findings on BioGPT. 
CURA maintains competitive discriminative performance across all five tasks, often matching or slightly surpassing the baseline. 
More importantly, our method demonstrates substantial advantages in calibration. 
While Deep Ensembles and MC Dropout struggle to improve calibration metrics over the baseline on the 7-day mortality task (with Brier scores remaining around 0.070), CURA significantly reduces the Brier score to 0.056 and the NLL from 0.246 to 0.187. 
This indicates that our method effectively handles uncertainty even in architectures different from that used in our primary analysis.

The results for ClinicalBERT, shown in Table~\ref{tab:clinicalbert_results}, further reinforce these observations. 
In the 30-day mortality prediction task, while existing uncertainty quantification methods provide marginal gains in discrimination, their impact on calibration is limited. 
In contrast, our method reduces the NLL from 0.334 (Baseline) to 0.289, achieving the best calibration performance among all compared methods. 
Furthermore, in the Early Discharge task, CURA achieves by far the largest reduction in AURC (from 0.082 to 0.025), whereas the other methods provide only modest improvements or even degrade performance.

Collectively, these additional experiments confirm that the benefits of our bi-level uncertainty fine-tuning objective are model-agnostic. 
The framework consistently reshapes the probability distributions to align with risks, delivering reliable uncertainty estimates without compromising the discriminative power of the underlying clinical language models.


\section{Implementation Details}
\label{app:experiment-setup}

\subsection{Clinical LM Fine-Tuning}
For all experiments, we train a separate task-specific model for each outcome and adopt a five-fold cross-validation scheme.
We train on the training split for $5$ epochs with a learning rate of $1\times10^{-5}$, for all models and all tasks without early stopping.

\subsection{Uncertainty Fine-Tuning and Baselines}
For all methods, we adopt the same multi-head classifier architecture described in Section~\ref{sec:base-loss} on top of the frozen embeddings, and use the same hyperparameters during training. 
We train for up to 50 epochs with a learning rate of $1 \times 10^{-4}$, applying early stopping based on validation negative log-likelihood (NLL).
The methods differ only in the training objective and in how uncertainty scores are derived from the heads, so that performance differences primarily reflect the uncertainty objective rather than model capacity.

\paragraph{Uncertainty Fine-Tuning} 
In Section~\ref{sec:uncertainty-fine-tuning}, we attach a 32-head classifier on top of the frozen embeddings and optimize it with the full uncertainty-aware objective $L_{total}$ in Section~\ref{sec:cohort}. 
Each head is implemented as a lightweight MLP that shares the same input features but maintains independent parameters.

We set the individual calibration weight to $\lambda_{ind} = 0.5$ and the cohort-alignment weight to $\lambda_{coh} = 0.01$ for all experiments. When computing the neighborhood risk $q(x)$ in Equation~\ref{eq:q-cohort}, we retrieve $K$ nearest neighbors using cosine distance. 
Specifically, we use $K=200$ nearest neighbors for the 7-day and 30-day mortality tasks, and $K=100$ neighbors for the remaining three tasks, reflecting their smaller cohort sizes.
Neighborhoods $\mathcal{N}_K(e_i)$ are pre-computed once using frozen embeddings to avoid computational overhead during uncertainty training. 
To prevent information leakage, the query sample itself is excluded from its neighbor set $\mathcal{N}_K(e_i)$.

\paragraph{MC Dropout} 
We train a single multi-head prediction network per task on top of the frozen embeddings. At inference time, we keep dropout active with a probability of $p=0.5$ and perform $T=10$ stochastic forward passes. The final prediction is obtained by averaging the predictive probabilities across these passes for each example.

\paragraph{Deep Ensembles} 
We train an ensemble of $M=5$ independently initialized classifier heads for each task, each using the same multi-head architecture as in Section~\ref{sec:base-loss}.
At test time, we average the output probabilities from all ensemble members.

\subsection{Uncertainty in Analyses}
Across all methods, we compute a scalar uncertainty $u(x)$ as the normalized binary entropy of the final predicted positive-class probability.
In practice, the binary entropy is computed using the natural logarithm, so the normalization constant $H_{\max}$ in Section~\ref{sec:calibration} corresponds to the maximum entropy $\log 2$ for a Bernoulli distribution.
Using this unified definition allows us to compare triage behavior and risk–coverage curves under a common uncertainty scale, rather than confounding the results with method-specific uncertainty proxies.

\begin{table*}[t!]
\centering
\small
\setlength{\tabcolsep}{5.5pt}
\renewcommand{\arraystretch}{1.2}
\begin{tabular}{llccccc}
\toprule
\textbf{Task} & \textbf{Method} & \textbf{AUROC} $\uparrow$ & \textbf{AUPRC} $\uparrow$ & \textbf{Brier} $\downarrow$ & \textbf{NLL} $\downarrow$ & \textbf{AURC} $\downarrow$ \\
\midrule
\multirow{4}{*}{\textbf{7-day Mortality}} 
& Baseline & 0.863 (0.011) & 0.080 (0.011) & 0.068 (0.009) & 0.246 (0.029) & 0.030 (0.023) \\
& Deep Ensemble & 0.863 (0.009) & 0.081 (0.007) & 0.070 (0.010) & 0.242 (0.027) & 0.025 (0.004) \\
& MC Dropout & \textbf{0.864} (0.009) & 0.081 (0.008) & 0.070 (0.010) & 0.250 (0.024) & 0.025 (0.009) \\
& \textbf{CURA (Ours)} & 0.863 (0.011) & \textbf{0.082} (0.008) & \textbf{0.056} (0.011) & \textbf{0.187} (0.023) & \textbf{0.010} (0.003) \\
\midrule
\multirow{4}{*}{\textbf{30-day Mortality}} 
& Baseline & \textbf{0.876} (0.005) & \textbf{0.228} (0.009) & 0.101 (0.003) & 0.316 (0.006) & 0.033 (0.002) \\
& Deep Ensemble & 0.874 (0.005) & 0.226 (0.011) & 0.097 (0.002) & 0.304 (0.003) & 0.031 (0.001) \\
& MC Dropout & \textbf{0.876} (0.005) & \textbf{0.228} (0.009) & 0.101 (0.003) & 0.315 (0.007) & 0.033 (0.002) \\
& \textbf{CURA (Ours)} & \textbf{0.876} (0.005) & 0.226 (0.007) & \textbf{0.088} (0.003) & \textbf{0.267} (0.005) & \textbf{0.021} (0.001) \\
\midrule
\multirow{4}{*}{\textbf{In-hospital Mortality}} 
& Baseline & 0.675 (0.015) & \textbf{0.033} (0.004) & 0.138 (0.005) & 0.422 (0.011) & 0.076 (0.005) \\
& Deep Ensemble & 0.671 (0.016) & \textbf{0.033} (0.004) & 0.131 (0.006) & 0.404 (0.015) & 0.070 (0.006) \\
& MC Dropout & 0.676 (0.015) & \textbf{0.033} (0.004) & 0.139 (0.006) & 0.426 (0.014) & 0.077 (0.006) \\
& \textbf{CURA (Ours)} & \textbf{0.678} (0.013) & \textbf{0.033} (0.003) & \textbf{0.116} (0.007) & \textbf{0.348} (0.017) & \textbf{0.046} (0.005) \\
\midrule
\multirow{4}{*}{\textbf{ICU Stay - 1}} 
& Baseline & 0.624 (0.006) & 0.044 (0.003) & 0.150 (0.008) & 0.470 (0.017) & 0.128 (0.009) \\
& Deep Ensemble & 0.621 (0.007) & 0.043 (0.004) & 0.140 (0.009) & 0.445 (0.020) & 0.109 (0.006) \\
& MC Dropout & 0.626 (0.008) & \textbf{0.045} (0.004) & 0.151 (0.009) & 0.473 (0.018) & 0.129 (0.006) \\
& \textbf{CURA (Ours)} & \textbf{0.630} (0.006) & 0.044 (0.003) & \textbf{0.123} (0.009) & \textbf{0.375} (0.019) & \textbf{0.057} (0.006) \\
\midrule
\multirow{4}{*}{\textbf{Early Discharge - 12}} 
& Baseline & 0.655 (0.014) & 0.066 (0.019) & 0.120 (0.020) & 0.406 (0.053) & 0.095 (0.021) \\
& Deep Ensemble & 0.654 (0.014) & 0.066 (0.014) & 0.111 (0.023) & 0.381 (0.063) & 0.082 (0.023) \\
& MC Dropout & 0.655 (0.015) & 0.063 (0.016) & 0.130 (0.022) & 0.431 (0.056) & 0.112 (0.027) \\
& \textbf{CURA (Ours)} & \textbf{0.656} (0.016) & \textbf{0.068} (0.013) & \textbf{0.100} (0.024) & \textbf{0.317} (0.064) & \textbf{0.042} (0.017) \\
\bottomrule
\end{tabular}
\caption{Performance of \textbf{BioClinicalBERT} on five clinical risk prediction tasks. Results are reported as mean (standard deviation) across five-fold cross-validation. 
``ICU Stay - 1'' refers to ICU stays longer than one day, and ``Early Discharge - 12'' refers to early discharge within 12 hours.}
\label{tab:bioclinicalbert_results}
\end{table*}

\begin{table*}[t!]
\centering
\small
\setlength{\tabcolsep}{5.5pt}
\renewcommand{\arraystretch}{1.2}
\begin{tabular}{llccccc}
\toprule
\textbf{Task} & \textbf{Method} & \textbf{AUROC} $\uparrow$ & \textbf{AUPRC} $\uparrow$ & \textbf{Brier} $\downarrow$ & \textbf{NLL} $\downarrow$ & \textbf{AURC} $\downarrow$ \\
\midrule
\multirow{4}{*}{\textbf{30-day Mortality}} 
& Baseline & \textbf{0.876} (0.003) & 0.228 (0.013) & 0.111 (0.001) & 0.334 (0.004) & 0.034 (0.001) \\
& Deep Ensemble & \textbf{0.876} (0.003) & \textbf{0.230} (0.012) & 0.109 (0.001) & 0.327 (0.003) & 0.034 (0.001) \\
& MC Dropout & \textbf{0.876} (0.003) & 0.229 (0.012) & 0.111 (0.001) & 0.333 (0.003) & 0.035 (0.001) \\
& \textbf{CURA (Ours)} & \textbf{0.876} (0.003) & 0.228 (0.013) & \textbf{0.098} (0.002) & \textbf{0.289} (0.002) & \textbf{0.024} (0.001) \\
\midrule
\multirow{4}{*}{\textbf{7-day Mortality}} 
& Baseline & 0.859 (0.010) & 0.080 (0.007) & 0.073 (0.005) & 0.241 (0.016) & 0.018 (0.003) \\
& Deep Ensemble & 0.859 (0.008) & \textbf{0.083} (0.009) & 0.071 (0.004) & 0.234 (0.010) & 0.017 (0.002) \\
& MC Dropout & 0.859 (0.008) & 0.082 (0.009) & 0.073 (0.005) & 0.241 (0.015) & 0.019 (0.003) \\
& \textbf{CURA (Ours)} & \textbf{0.860} (0.008) & 0.077 (0.006) & \textbf{0.064} (0.003) & \textbf{0.196} (0.007) & \textbf{0.010} (0.001) \\
\midrule
\multirow{4}{*}{\textbf{In-hospital Mortality}} 
& Baseline & \textbf{0.685} (0.012) & \textbf{0.034} (0.002) & 0.169 (0.005) & 0.478 (0.010) & 0.097 (0.005) \\
& Deep Ensemble & 0.684 (0.012) & \textbf{0.034} (0.002) & 0.165 (0.007) & 0.466 (0.016) & 0.093 (0.008) \\
& MC Dropout & \textbf{0.685} (0.012) & \textbf{0.034} (0.002) & 0.170 (0.005) & 0.480 (0.010) & 0.098 (0.005) \\
& \textbf{CURA (Ours)} & 0.684 (0.012) & \textbf{0.034} (0.002) & \textbf{0.150} (0.006) & \textbf{0.419} (0.013) & \textbf{0.071} (0.006) \\
\midrule
\multirow{4}{*}{\textbf{ICU Stay - 1}} 
& Baseline & 0.645 (0.013) & 0.045 (0.004) & 0.188 (0.006) & 0.551 (0.013) & 0.157 (0.013) \\
& Deep Ensemble & 0.643 (0.011) & \textbf{0.046} (0.004) & 0.182 (0.004) & 0.535 (0.012) & 0.144 (0.011) \\
& MC Dropout & 0.645 (0.013) & 0.045 (0.003) & 0.190 (0.006) & 0.558 (0.015) & 0.162 (0.016) \\
& \textbf{CURA (Ours)} & \textbf{0.647} (0.013) & \textbf{0.046} (0.004) & \textbf{0.170} (0.010) & \textbf{0.483} (0.023) & \textbf{0.109} (0.013) \\
\midrule
\multirow{4}{*}{\textbf{Early Discharge - 12}} 
& Baseline & 0.668 (0.014) & 0.067 (0.017) & 0.103 (0.007) & 0.353 (0.018) & 0.082 (0.015) \\
& Deep Ensemble & 0.666 (0.014) & \textbf{0.071} (0.016) & 0.092 (0.006) & 0.320 (0.015) & 0.062 (0.006) \\
& MC Dropout & 0.667 (0.015) & 0.068 (0.016) & 0.107 (0.007) & 0.364 (0.018) & 0.085 (0.013) \\
& \textbf{CURA (Ours)} & \textbf{0.670} (0.014) & 0.056 (0.012) & \textbf{0.082} (0.005) & \textbf{0.262} (0.017) & \textbf{0.025} (0.004) \\
\bottomrule
\end{tabular}
\caption{Performance of \textbf{ClinicalBERT} on five clinical risk prediction tasks. Results are reported as mean (standard deviation) across five-fold cross-validation. 
``ICU Stay - 1'' refers to ICU stays longer than one day, and ``Early Discharge - 12'' refers to early discharge within 12 hours.}
\label{tab:clinicalbert_results}
\end{table*}

\section{Additional Ablation Results}
\label{app:ablation_full}
In Section~\ref{sec:ablation}, we analyzed the contributions of our uncertainty-aware components on two high-stakes tasks. 
Here, we extend this analysis to the remaining three outcomes on BioGPT: 30-day mortality, ICU stay longer than one day, and early discharge within 12 hours. 
The detailed results are reported in Table~\ref{tab:ablation_full}.

These results corroborate the findings presented in the main text. 
We observe that introducing either the individual calibration term $L_{ind}$ or the cohort-aware alignment term $L_{coh}$ consistently reduces calibration error (Brier score, NLL, and AURC) compared to the baseline.
Notably, for the 30-day mortality task, $L_{ind}$ alone drives substantial gains in calibration while simultaneously boosting discriminative performance. 
Across all three tasks, the full CURA objective—combining both terms—achieves the best overall calibration performance. Furthermore, this improvement in reliability does not come at the cost of accuracy; CURA maintains, and in most cases slightly improves, the AUROC and AUPRC scores relative to the baseline.


\section{Sensitivity to Loss Weights}
\label{app:loss_weight_sensitivity}

The relative strengths of the individual- and cohort-level regularizers are controlled by $\lambda_{ind}$ in Equation~\ref{eq:L-indiv} and $\lambda_{coh}$ through $w(x_i)$ in Equation~\ref{eq:L-coh}. 
To evaluate the robustness of these choices, we conduct a sensitivity analysis on BioGPT for the two high-stakes mortality tasks: 7-day mortality and in-hospital mortality. 
We first vary $\lambda_{ind}$ while fixing $\lambda_{coh}=0.01$, and then vary $\lambda_{coh}$ while fixing $\lambda_{ind}=0.5$. 
All other training and evaluation settings are identical to the main experiments.

Tables~\ref{tab:lambda_sensitivity_7day} and~\ref{tab:lambda_sensitivity_inhospital} show that $\lambda_{ind}$ around 0.5 gives a strong balance between discrimination and calibration on both tasks. 
Increasing $\lambda_{ind}$ beyond 0.5 yields only modest additional calibration gains and can slightly hurt AUROC/AUPRC. 
For $\lambda_{coh}$, small positive values in the range 0.01--0.05 consistently improve calibration metrics, whereas larger values offer no clear further benefit. 
Overall, these results indicate that our default setting $(\lambda_{ind}=0.5,\lambda_{coh}=0.01)$ lies in a stable region rather than being finely tuned to a single coefficient choice.

\begin{table}[t!]
\centering
\small
\setlength{\tabcolsep}{3.5pt}
\renewcommand{\arraystretch}{1.2}
\begin{tabular}{lccccc}
\toprule
\textbf{Method} & \textbf{AUROC} & \textbf{AUPRC} & \textbf{Brier} & \textbf{NLL} & \textbf{AURC} \\
\midrule
\multicolumn{6}{l}{\textit{30-day Mortality}} \\
Baseline & 0.881 & 0.270 & 0.064 & 0.231 & 0.024 \\
w/ $L_{\mathrm{ind}}$ & \textbf{0.890} & \textbf{0.280} & 0.045 & 0.148 & 0.012 \\
w/ $L_{\mathrm{coh}}$ & 0.881 & 0.268 & 0.053 & 0.212 & 0.023 \\
\textbf{CURA} & \textbf{0.890} & \textbf{0.280} & \textbf{0.038} & \textbf{0.146} & \textbf{0.009} \\
\midrule
\multicolumn{6}{l}{\textit{ICU Stay - 1}} \\
Baseline & 0.579 & 0.035 & 0.105 & 0.350 & 0.064 \\
w/ $L_{\mathrm{ind}}$ & \textbf{0.584} & \textbf{0.035} & 0.086 & 0.295 & 0.049 \\
w/ $L_{\mathrm{coh}}$ & 0.579 & 0.035 & 0.094 & 0.330 & 0.063 \\
\textbf{CURA} & \textbf{0.584} & \textbf{0.035} & \textbf{0.085} & \textbf{0.288} & \textbf{0.040} \\
\midrule
\multicolumn{6}{l}{\textit{Early Discharge - 12}} \\
Baseline & 0.587 & 0.027 & 0.018 & 0.085 & 0.007 \\
w/ $L_{\mathrm{ind}}$ & \textbf{0.594} & \textbf{0.031} & 0.013 & 0.067 & \textbf{0.006} \\
w/ $L_{\mathrm{coh}}$ & 0.587 & 0.028 & 0.016 & 0.076 & \textbf{0.006} \\
\textbf{CURA} & \textbf{0.594} & \textbf{0.031} & \textbf{0.010} & \textbf{0.056} & \textbf{0.006} \\
\bottomrule
\end{tabular}
\caption{Full ablation study results on \textbf{BioGPT} for the remaining clinical tasks. We isolate the contributions of individual calibration $L_{ind}$ and cohort alignment $L_{coh}$. Standard deviations are omitted for brevity.}
\label{tab:ablation_full}
\end{table}

\begin{table*}[t]
\centering
\small
\setlength{\tabcolsep}{5pt}
\renewcommand{\arraystretch}{1.15}
\begin{tabular}{lccccc}
\toprule
\textbf{Method / Setting} & \textbf{AUROC}$\uparrow$ & \textbf{AUPRC}$\uparrow$ & \textbf{Brier}$\downarrow$ & \textbf{NLL}$\downarrow$ & \textbf{AURC}$\downarrow$ \\
\midrule
Baseline & 0.852 & 0.127 & 0.032 & 0.120 & 0.008 \\
\midrule
\multicolumn{6}{l}{\textit{Vary $\lambda_{ind}$ ($\lambda_{coh}=0.01$)}} \\
$\lambda_{ind}=0.05$ & 0.859 & 0.127 & 0.032 & 0.121 & 0.007 \\
$\lambda_{ind}=0.10$ & 0.887 & 0.150 & 0.030 & 0.127 & 0.006 \\
$\lambda_{ind}=0.20$ & 0.891 & 0.157 & 0.025 & 0.110 & 0.004 \\
$\lambda_{ind}=0.50$ & 0.892 & 0.160 & 0.015 & 0.075 & 0.002 \\
$\lambda_{ind}=1.00$ & 0.890 & 0.159 & 0.012 & 0.075 & 0.002 \\
\midrule
\multicolumn{6}{l}{\textit{Vary $\lambda_{coh}$ ($\lambda_{ind}=0.5$)}} \\
$\lambda_{coh}=0.005$ & 0.892 & 0.160 & 0.016 & 0.081 & 0.003 \\
$\lambda_{coh}=0.010$ & 0.892 & 0.160 & 0.015 & 0.075 & 0.002 \\
$\lambda_{coh}=0.050$ & 0.893 & 0.161 & 0.013 & 0.072 & 0.001 \\
$\lambda_{coh}=0.100$ & 0.892 & 0.160 & 0.015 & 0.076 & 0.002 \\
$\lambda_{coh}=0.500$ & 0.892 & 0.161 & 0.016 & 0.078 & 0.003 \\
\bottomrule
\end{tabular}
\caption{Sensitivity of CURA to $\lambda_{ind}$ and $\lambda_{coh}$ on BioGPT for 7-day mortality. All other training and evaluation settings are identical to the main experiment. Standard deviations are omitted for brevity.}
\label{tab:lambda_sensitivity_7day}
\end{table*}

\begin{table*}[t]
\centering
\small
\setlength{\tabcolsep}{5pt}
\renewcommand{\arraystretch}{1.15}
\begin{tabular}{lccccc}
\toprule
\textbf{Method / Setting} & \textbf{AUROC}$\uparrow$ & \textbf{AUPRC}$\uparrow$ & \textbf{Brier}$\downarrow$ & \textbf{NLL}$\downarrow$ & \textbf{AURC}$\downarrow$ \\
\midrule
Baseline & 0.621 & 0.027 & 0.044 & 0.175 & 0.015 \\
\midrule
\multicolumn{6}{l}{\textit{Vary $\lambda_{ind}$ ($\lambda_{coh}=0.01$)}} \\
$\lambda_{ind}=0.05$ & 0.630 & 0.028 & 0.042 & 0.169 & 0.014 \\
$\lambda_{ind}=0.10$ & 0.638 & 0.029 & 0.042 & 0.164 & 0.013 \\
$\lambda_{ind}=0.20$ & 0.642 & 0.029 & 0.038 & 0.150 & 0.012 \\
$\lambda_{ind}=0.50$ & 0.641 & 0.029 & 0.029 & 0.124 & 0.011 \\
$\lambda_{ind}=1.00$ & 0.639 & 0.028 & 0.023 & 0.113 & 0.010 \\
\midrule
\multicolumn{6}{l}{\textit{Vary $\lambda_{coh}$ ($\lambda_{ind}=0.5$)}} \\
$\lambda_{coh}=0.005$ & 0.642 & 0.029 & 0.031 & 0.129 & 0.011 \\
$\lambda_{coh}=0.010$ & 0.641 & 0.029 & 0.029 & 0.124 & 0.011 \\
$\lambda_{coh}=0.050$ & 0.642 & 0.029 & 0.026 & 0.116 & 0.009 \\
$\lambda_{coh}=0.100$ & 0.641 & 0.029 & 0.031 & 0.125 & 0.012 \\
$\lambda_{coh}=0.500$ & 0.641 & 0.029 & 0.033 & 0.128 & 0.013 \\
\bottomrule
\end{tabular}
\caption{Sensitivity of CURA to $\lambda_{ind}$ and $\lambda_{coh}$ on BioGPT for in-hospital mortality. All other training and evaluation settings are identical to the main experiment. Standard deviations are omitted for brevity.}
\label{tab:lambda_sensitivity_inhospital}
\end{table*}

\section{Cohort-Aware Loss as Soft-Label Cross-Entropy}
\label{app:cohort-softlabel}

For completeness, we show that the combination of the base risk loss $L_{\text{base}}$ and the cohort-aware loss $L_{\text{coh}}$ can be equivalently written as a single cross-entropy with a cohort-informed soft label.

Consider a single example $x$ with binary label $y\in\{0,1\}$, cohort risk $q(x)\in[0,1]$, and ensemble-averaged prediction $\bar p(x)\in(0,1)$. Denote the binary cross-entropy between a prediction $p$ and target $r$ by
\begin{equation}
\label{eq:bce_def}
\begin{aligned}
    \mathrm{CE}(p,r) &= -\bigl[r\log p + (1-r)\log(1-p)\bigr].
\end{aligned}
\end{equation}

The per-example contributions of the base loss and the cohort-aware loss are
\begin{equation}
\label{eq:per_example_loss}
\begin{aligned}
    \ell_{\text{base}}(x) &= \mathrm{CE}\bigl(\bar p(x), y\bigr),\\
    \ell_{\text{coh}}(x) &= w(x)\,\mathrm{CE}\bigl(\bar p(x), q(x)\bigr),
\end{aligned}
\end{equation}
where $w(x) = \lambda_{\text{coh}}\,\hat H\bigl(q(x)\bigr)\ge 0$. Their sum can be written as
\begin{equation}
\label{eq:loss_sum_expanded}
\begin{aligned}
    &\ell_{\text{base}}(x) + \ell_{\text{coh}}(x) = \\
    &-\Bigl[y\log\bar p(x) + \bigl(1-y\bigr)\log\bigl(1-\bar p(x)\bigr)\Bigr] \\
      &- w(x)\Bigl[q(x)\log\bar p(x) \\
      &+ \bigl(1-q(x)\bigr)\log\bigl(1-\bar p(x)\bigr)\Bigr].
\end{aligned}
\end{equation}

Grouping the coefficients of $\log\bar p(x)$ and $\log\bigl(1-\bar p(x)\bigr)$ yields
\begin{equation}
\label{eq:loss_grouped}
\begin{aligned}
    &\ell_{\text{base}}(x) + \ell_{\text{coh}}(x) = \\
     &-\Bigl[\bigl(y + w(x)q(x)\bigr)\log\bar p(x) \\
    &+ \bigl(1-y + w(x)(1-q(x))\bigr)\log\bigl(1-\bar p(x)\bigr)\Bigr].
\end{aligned}
\end{equation}

Define the mixing weight
\begin{equation}
\label{eq:mixing_weight}
\begin{aligned}
    \gamma(x) &= \frac{w(x)}{1+w(x)}\in[0,1),
\end{aligned}
\end{equation}
and the cohort-aware soft label
\begin{equation}
\label{eq:soft_label}
\begin{aligned}
    t(x) &= (1-\gamma(x))\,y + \gamma(x)\,q(x) \\
         &= \frac{y + w(x)q(x)}{1+w(x)}.
\end{aligned}
\end{equation}

Using $1-t(x) = \frac{1-y + w(x)(1-q(x))}{1+w(x)}$, we can rewrite the sum as
\begin{equation}
\label{eq:loss_rewritten}
\begin{aligned}
    \ell_{\text{base}}(x)+&\ell_{\text{coh}}(x) = (1+w(x))\,\mathrm{CE}\bigl(\bar p(x), t(x)\bigr).
\end{aligned}
\end{equation}

Taking expectations over $(x,y)$ gives
\begin{equation}
\label{eq:final_objective}
\begin{aligned}
    L_{\text{base}} + L_{\text{coh}} &= \mathbb{E}_{(x,y)}\bigl[(1+w(x))\,\mathrm{CE}\bigl(\bar p(x), t(x)\bigr)\bigr],
\end{aligned}
\end{equation}
showing that our objective is equivalent to optimizing a cross-entropy loss with a cohort-informed soft label $t(x)$ and a sample-dependent weight $1+w(x)$. This interpretation highlights $L_{\text{coh}}$ as a form of neighborhood-informed label smoothing that focuses more strongly on ambiguous patient cohorts.

\begin{figure*}[t]
    \centering
    \includegraphics[width=\textwidth]{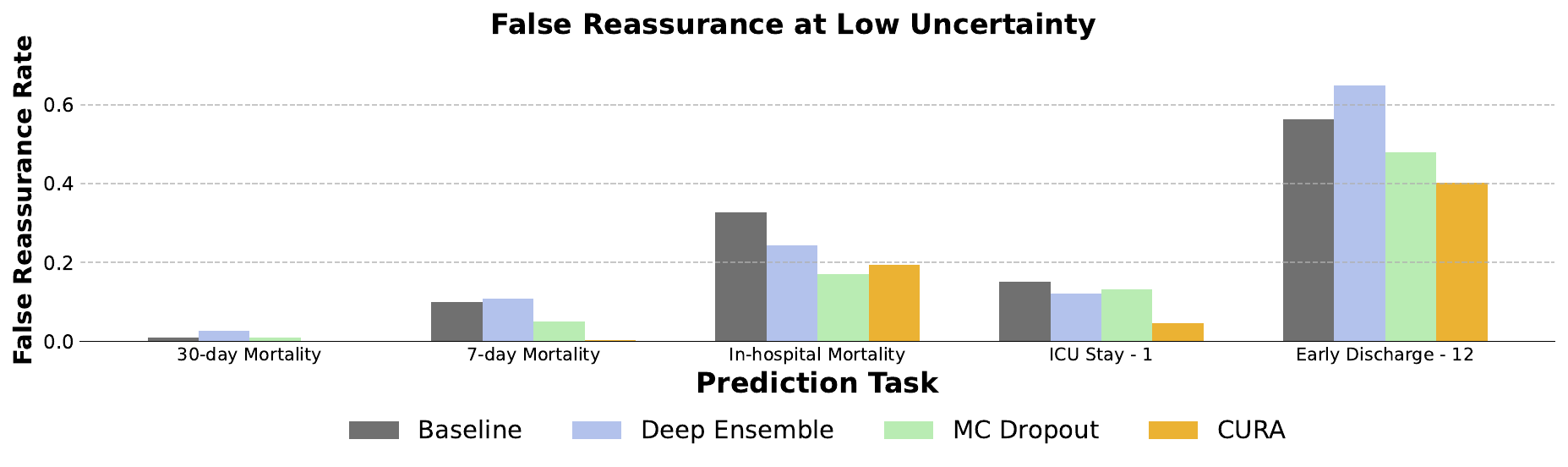}
    \caption{
    False reassurance rates ($\tau=0.05$) at low uncertainty across five prediction tasks with BioGPT. 
    Lower values indicate fewer high-risk patients being confidently misclassified as safe.
    }
    \label{fig:false_reassurance_0.05}
\end{figure*}

\begin{figure*}[t]
    \centering
    \includegraphics[width=\textwidth]{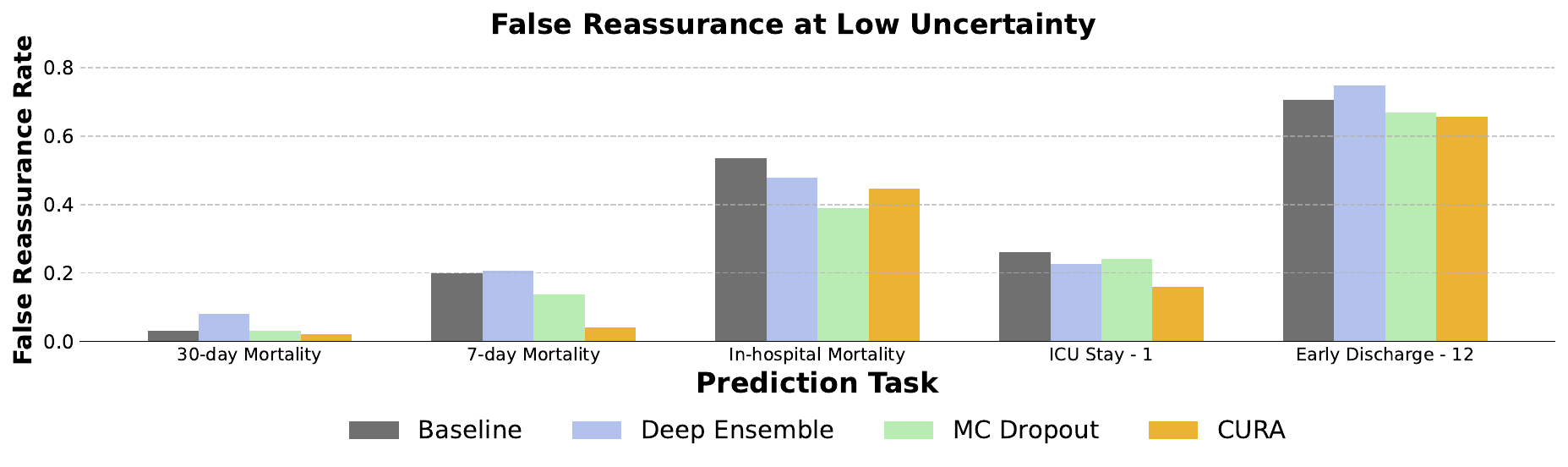}
    \caption{
    False reassurance rates ($\tau=0.15$) at low uncertainty across five prediction tasks with BioGPT. 
    Lower values indicate fewer high-risk patients being confidently misclassified as safe.
    }
    \label{fig:false_reassurance_0.15}
\end{figure*}

\section{Definition of False Reassurance Rate}
\label{app:false-reassurance}


In the diagnostic testing literature, the \emph{false reassurance rate} is commonly used to quantify the risk of missed disease among patients who receive a negative test result. Classical definitions take the false reassurance rate to be the complement of the negative predictive value, that is
\[
\text{FRR} = 1 - \text{NPV} = \frac{\text{false negatives}}{\text{false negatives} + \text{true negatives}},
\]
representing the proportion of diseased individuals among all test-negative patients~\citep{mitchell2005frozen}.
This notion captures an important failure mode in screening programmes: patients who are incorrectly reassured by a negative result.

In our setting, models output both a predicted risk score $p(x)$ and an uncertainty score $u(x)$ for each patient $x$. We are specifically concerned with the subset of high-risk patients who are assigned to a region that appears \emph{both} low-risk and highly certain. To formalize this, we define a safe region using thresholds $\tau$:
\[
\mathcal{S} = \{x : u(x) < \tau,\ p(x) < \tau \}.
\]
We then define the task-specific \emph{False Reassurance Rate (FRR)} as
\begin{equation}
\text{FRR} =
\frac{\big|\{x : y = 1,\ x \in \mathcal{S}\}\big|}
     {\big|\{x : y = 1\}\big|},
\end{equation}
where $y = 1$ denotes that the patient experiences the adverse outcome, and $|\cdot|$ denotes set cardinality. 
Intuitively, FRR measures the fraction of truly positive patients whose predictions fall into the low-uncertainty, low-risk region $\mathcal{S}$, i.e., high-risk patients who are confidently misclassified as safe.

Our definition is conceptually related to the classical false reassurance rate, but differs in two aspects. First, the denominator conditions on all positive cases rather than all test-negative cases, reflecting the perspective of ``among high-risk patients, how many are given misleading reassurance?''. Second, by restricting to the safe region $\mathcal{S}$, we only count those false negatives for which the model is simultaneously low-risk and low-uncertainty, which are the most concerning from a triage standpoint. 
In the experiments reported in Section~\ref{sec:fa-false-reassurance}, we set $\tau = 0.1$ for all tasks.

\section{Sensitivity Analysis of False Reassurance Thresholds}
\label{app:false-reassurance-results}
In Section~\ref{sec:fa-false-reassurance}, we presented False Reassurance Rates (FRR) using a fixed safety threshold of $\tau = 0.1$ for both uncertainty and predicted risk. 
To evaluate the robustness of CURA against this hyperparameter choice, we report FRR results using stricter ($\tau = 0.05$) and looser ($\tau = 0.15$) thresholds in Figure~\ref{fig:false_reassurance_0.05} and Figure~\ref{fig:false_reassurance_0.15}, respectively.
Results across both alternative thresholds exhibit patterns consistent with the main-text findings at $\tau = 0.1$. Tightening the threshold to $\tau = 0.05$ constrains the safe region and lowers FRR for all methods; however, CURA generally maintains the lowest or near-lowest false reassurance rates on the acute mortality and early-discharge tasks, while remaining competitive with top baselines on in-hospital mortality and ICU stay prediction. Conversely, relaxing the threshold to $\tau = 0.15$ universally increases FRR, yet CURA continues to yield fewer confidently misclassified high-risk patients than alternative approaches, demonstrating particularly distinct gains on the 7-day mortality, 30-day mortality, and early-discharge tasks. Taken together with Figure~\ref{fig:false_reassurance}, these sensitivity analyses indicate that CURA’s reduction in false reassurance is not dependent on a finely tuned choice of $\tau$, and its safety benefits persist under both stricter and more permissive operating points of the BioGPT backbone.

\section{Runtime Comparison}
\label{app:runtime}
To assess the practical overhead of CURA, we compare the total uncertainty fine-tuning runtime of different methods under identical hardware, classifier architecture, and early-stopping settings on the 7-day mortality task with the BioGPT backbone. 
All methods use the same extracted embeddings and the same multi-head classifier, so the comparison isolates the overhead introduced by the uncertainty mechanism itself. 
For CURA, the $K$-nearest-neighbor cohorts are pre-computed once at the beginning of each run, and this pre-computation time is included in the reported GPU hours.

As shown in Table~\ref{tab:runtime_comparison}, CURA has runtime comparable to, and in our setting slightly shorter than, the single-model baseline, while remaining substantially more efficient than Deep Ensembles. 
This result supports our claim that CURA functions as a lightweight plug-in objective: it adds two loss terms computed from existing probabilities and pre-computed neighborhoods, without requiring multiple independently trained backbones or repeated stochastic inference,
resonating with broader trends toward lightweight or data-efficient post-training objectives, including uncertainty-aware fine-tuning, weak supervision, active-learning-based reward modeling, and balanced preference optimization~\citep{krishnan2024enhancing,tong2024optimizing,wang2025bpo,duan2025efficient}.

\begin{table}[t]
\centering
\small
\setlength{\tabcolsep}{8pt}
\renewcommand{\arraystretch}{1.15}
\begin{tabular}{lc}
\toprule
\textbf{Method} & \textbf{Time / GPU hours} \\
\midrule
Baseline & 1.03 \\
Deep Ensemble & 4.43 \\
MC Dropout & 1.08 \\
CURA (Ours) & \textbf{0.94} \\
\bottomrule
\end{tabular}
\caption{Runtime comparison of uncertainty fine-tuning methods on the 7-day mortality task with the BioGPT backbone. 
Values denote total GPU hours on a single NVIDIA A40 GPU under identical hardware, classifier architecture, and early-stopping settings, aggregated over five-fold cross-validation. 
For CURA, the one-time pre-computation of $K$-nearest-neighbor cohorts is included in the reported runtime.}
\label{tab:runtime_comparison}
\end{table}

\section{Computing Infrastructure and Resources}
\label{app:compute}

\subsection{Model Sizes}
We experiment with three backbone language models. BERT-based models such as ClinicalBERT and BioClinicalBERT contain approximately 110 million parameters, while the GPT-style BioGPT backbone contains about 347 million parameters.

\subsection{Hardware and Compute Environment}
Most experiments were conducted on A40 GPUs with 48 GB of VRAM provided by Research Infrastructure Services compute1 cluster, which is a secure service approved for ePHI, protected, or confidential data.

\section{Use of AI Assistants}
AI assistants were used only for grammar and syntax revision of the manuscript. 
All experiments, analyses, and substantive writing were conceived and produced by the authors without AI-generated content.

\end{document}